\documentclass[final,3p]{elsarticle}

\usepackage{lineno}
\usepackage[colorlinks=true,breaklinks=true,pdftex]{hyperref}
\usepackage{amsmath,amsfonts}
\usepackage{graphicx}
\usepackage{wrapfig}
\usepackage{multirow}
\modulolinenumbers[1]

\journal{GMP 2024}

\biboptions{authoryear}

\begin{document}

\begin{frontmatter}

\title{PointeNet: A Lightweight Framework for Effective and Efficient Point Cloud Analysis}


\author[first]{Lipeng Gu}
\ead{gulp1224@nuaa.edu.cn}
\author[first]{Xuefeng Yan}
\ead{yxf@nuaa.edu.cn}
\author[second]{Liangliang Nan}
\ead{Liangliang.Nan@tudelft.nl}
\author[third]{Dingkun Zhu}
\ead{zhudingkun@jsut.edu.cn}
\author[first]{Honghua Chen}
\ead{chenhonghuacn@gmail.com}
\author[fourth]{Weiming Wang}
\ead{wmwang@hkmu.edu.hk}
\author[first]{Mingqiang Wei}
\ead{mingqiang.wei@gmail.com}
\address[first]{Nanjing University of Aeronautics and Astronautics}
\address[second]{Delft University of Technology}
\address[third]{Jiangsu University of Technology}
\address[fourth]{Hong Kong Metropolitan University}

\begin{abstract}
Current methodologies in point cloud analysis predominantly explore 3D geometries, often achieved through the introduction of intricate learnable geometric extractors in the encoder or by deepening networks with repeated blocks.
However, these approaches inevitably lead to a significant number of learnable parameters, resulting in substantial computational costs and imposing memory burdens on CPU/GPU. 
Additionally, the existing strategies are primarily tailored for object-level point cloud classification and segmentation tasks, with limited extensions to crucial scene-level applications, such as autonomous driving.
In response to these limitations, we introduce PointeNet, an \textbf{e}fficient \textbf{net}work designed specifically for \textbf{point} cloud analysis.
PointeNet distinguishes itself with its lightweight architecture, low training cost, and plug-and-play capability, effectively capturing representative features.
The network consists of a Multivariate Geometric Encoding (MGE) module and an \textit{optional} Distance-aware Semantic Enhancement (DSE) module. 
The MGE module employs operations of sampling, grouping, and multivariate geometric aggregation to lightweightly capture and adaptively aggregate multivariate geometric features, providing a comprehensive depiction of 3D geometries.
The DSE module, designed for real-world autonomous driving scenarios, enhances the semantic perception of point clouds, particularly for distant points. 
Our method demonstrates flexibility by seamlessly integrating with a classification/segmentation head or embedding into off-the-shelf 3D object detection networks, achieving notable performance improvements at a minimal cost.
Extensive experiments on object-level datasets, including ModelNet40, ScanObjectNN, ShapeNetPart, and the scene-level dataset KITTI, demonstrate the superior performance of PointeNet over state-of-the-art methods in point cloud analysis.
Notably, PointeNet outperforms PointMLP with significantly fewer parameters on ModelNet40, ScanObjectNN, and ShapeNetPart, and achieves a substantial improvement of nearly 3\% in $3D\ AP_{R40}$ for PointRCNN on KITTI with a minimal parameter cost of 1.4 million.
\end{abstract}


\end{frontmatter}


\section{Introduction}

With the popularity of 3D sensors such as LiDARs and depth cameras \cite{kitti}, point cloud analysis has gained significant prominence in both academic research and industrial development \cite{pointnet,pointnet++,votenet,bevfusion}.
Unlike grid-based RGB images, point clouds comprise unordered and irregular points that outline the surfaces of objects in 3D space.
This characteristic poses significant challenges in designing effective point cloud analysis models.

PointNet/PointNet++ \cite{pointnet,pointnet++} are pioneering works capable of directly analyzing unordered point clouds without the need for preprocessing.
Just as ResNet \cite{resnet} serves as a prominent backbone in image processing, PointNet/PointNet++ has emerged as a widely adopted backbone network in subsequent point cloud analysis models \cite{votenet,3DSSD,PointRCNN,pointmlp}.
The iterative evolution in these methods adheres to the ``\textit{continual increment}'' principle, primarily focusing on modifying the backbone network, PointNet++, to capture more representative local geometric features.
Specific improvements fall mainly into two aspects: introducing well-designed local geometric extractors (such as Graph Convolution \cite{dgcnn,3D-GCN}, Adaptive Point Convolution \cite{paconv}, and Residual Point Block \cite{pointmlp}) into the encoder or simply stacking repeated blocks to deepen the network \cite{pointmlp,RSCNN}.
Such strategies, while capable of boosting performance, inevitably lead to a notable surge in network parameters. 
Fortunately, Point-NN recognizes this longstanding challenge and for the first time attempts to employ a counterintuitive ``\textit{subtraction}'' strategy by trimming PointNet++, retaining only its non-learnable components—sampling, grouping, and pooling. This allows tasks like point cloud classification, segmentation, and even 3D detection to be performed without the need for training.
While this unique strategy significantly reduces training costs and achieves point cloud analysis, it falls short of exploiting the local geometry properties of point clouds, relying solely on spatial neighboring information.
Furthermore, existing methods, including Point-NN, primarily focus on object-level point cloud analysis and see a limited extension to more valuable scene-level applications, such as autonomous driving.
These problems hinder Point-NN from fully unleashing its potential.

In this paper, we propose an \textbf{e}fficient \textbf{Net}work for \textbf{\textit{point}} cloud analysis, dubbed \textbf{\textit{PointeNet}}.
Our model is highly streamlined, comprising only non-parametric computational components such as sampling, grouping, pooling, and a minimal number of learnable parameters, and can be flexibly combined with a point cloud classification/segmentation head, or be embedded into cutting-edge 3D detection networks tailored for real-world autonomous driving scenarios to enhance performance.
Specifically, PointeNet comprises two crucial modules: the multivariate geometric encoding (MGE) module, and the distance-aware semantic enhancement (DSE) module.
MGE comprises non-learnable farthest point sampling (FPS) and k-nearest neighbor (k-NN) components, along with a multivariate local geometric aggregation (MLGA) module incorporating a minimal number of learnable parameters.
This module captures multivariate 3D geometric features within local regions of point clouds, encompassing curvature, normal, and spatial neighboring information.
DSE is an optional module tailored for autonomous driving scenarios.
It dynamically adjusts the segmentation difficulty on a point-by-point basis, particularly allocating more ``attention'' to challenging distant points, and outputs distance-aware semantic features to rich point clouds, further enhancing the performance of arbitrary 3D detection networks.
We evaluate the performance of PointeNet through a comparative analysis with nineteen competitors across the ScanObjectNN \cite{ScanObjectNN}, ModelNet40 \cite{ModelNet40}, ShapeNetPart \cite{ShapeNetPart}, and KITTI \cite{kitti} datasets. Remarkably, PointeNet surpasses all competitors, showcasing its superior performance. The contributions of this work are threefold:
\begin{itemize}
    \item We analyze the longstanding challenge in recent point cloud analysis methods and propose an efficient network, dubbed PointeNet. It excels in lightweight capturing and adaptive aggregation of multivariate geometric and semantic features, making it suitable for point cloud segmentation/classification and enhancing arbitrary 3D object detection networks tailored for autonomous driving scenarios.
    \item We propose a multivariate geometric encoding (MGE), featuring non-learnable FPS and k-NN modules, complemented by a minimal number of learnable parameters, to capture diverse 3D geometric features.
    \item We propose an optional distance-aware semantic enhancement (DSE) module tailored for autonomous driving scenarios. It dynamically adjusts the segmentation difficulty on a point-by-point basis, giving more attention to challenging distant points, and outputs distance-aware semantic features.
\end{itemize}

\section{Related Work}

\subsection{Point Cloud Analysis}

There are two main paradigms for processing irregular and unordered point clouds: grid-based and point-based methods. 
Grid-based methods \cite{bevfusion,votenet,pv-rcnn,pointpillars} involve projecting irregular point clouds onto regular grids, such as pillars or voxels, and then processing them using 2D/3D convolutional neural networks (CNNs).
This paradigm significantly enhances the processing speed by transforming the point cloud into structured grids. 
However, projecting onto grids may lead to information loss, degrading the quality of the point cloud representation \cite{std}.

In contrast, point-based methods have emerged to directly process irregular point clouds without additional regularization, preserving the point cloud details more accurately. 
PointNet \cite{pointnet} is a pioneering method that utilizes shared MLPs to handle unordered point clouds as input directly. 
PointNet++ \cite{pointnet++} builds upon PointNet by introducing a hierarchical feature learning paradigm, allowing for the recursive capture of local geometric structures. 
Due to the promising performance exhibited by PointNet++, particularly in leveraging local geometric representations, including multi-scale geometric information, it has become a foundational element in modern point cloud analysis methods. 
Subsequent works, such as KPConv \cite{KPConv}, RSCNN \cite{RSCNN}, 3D-GCN \cite{3D-GCN}, PAConv \cite{paconv}, PointConv \cite{Pointconv} and PointMLP \cite{pointmlp}, either introduce intricate local geometric extractors or simply stack repeated network blocks to achieve further performance improvements. However, these strategies make the networks more complex and less efficient.
\textit{In contrast, our PointeNet draws from the successful experience of Point-NN and proposes an efficient network primarily based on non-learnable components, capturing representative multivariate geometric and semantic features at a minimal cost in terms of parameters. 
}

\subsection{Local Geometry Exploration}

Since the powerful ability to capture local geometric features of point clouds demonstrated by PointNet++ \cite{pointnet++}, subsequent research has mainly focused on exploring local geometric representations, which can be divided into three categories: convolution-based, graph-based, and attention-based methods.
Classic convolution-based methods, with PointConv \cite{Pointconv} as a representative example, employ MLPs to approximate continuous weights and density functions within convolution filters. It extends dynamic filtering to a novel convolution operation.
Unlike convolution-based methods, graph-based methods explore relationships between points. For instance, DGCNN \cite{dgcnn} introduces EdgeConv, a novel method that generates edge features describing the relationships between points and their neighbors, capturing local features of the point cloud.
3D-GCN \cite{3D-GCN} employs a 3D graph convolutional network to form deformable 3D kernels, directly performing convolutional computations on point clouds.
Related to graph-based methods, attention-based methods similarly focus on exploring direct relationships between points, as seen in works like PCT \cite{PCT} and Point Transformer \cite{PointTransformer}. 
Although these methods have achieved success in point cloud analysis, they have consistently employed an ``\textit{increment}'' principle, undoubtedly introducing more learnable parameters and thereby reducing computational efficiency and exacerbating the burden on CPU/GPU.
Compared with these methods, our PointeNet achieves better performance with fewer parameters.


\subsection{Deep Network Architecture for Point Clouds}

The advancement of methods in point cloud analysis often coincides with breakthroughs in image processing networks.
For instance, following the success of ResNet \cite{resnet} based on simple convolutions in the field of image processing, subsequent methods often use it as a backbone and continuously improve upon it. 
Similarly, after the breakthrough achieved by PointNet/PointNet++ based on simple MLPs in point cloud processing, they have become the mainstream backbones for point cloud analysis.
After the success of graph- \cite{felzenszwalb2004efficient}, attention- \cite{wang2018non-local}, and transformer-based \cite{dosovitskiy2020image} methods in the field of image processing, they have inspired a series of works in point cloud analysis, contributing to the further development of point cloud analysis methods \cite{Pointconv,dgcnn,ran2021learning,PCT,PointTransformer}.
Our PointeNet incorporates the strengths of previous methods, enabling the lightweight capturing and adaptive aggregation of diverse geometric and semantic representations of point clouds at a lower cost, thereby achieving powerful point cloud analysis capabilities. 

\section{Methodology}

\subsection{Overview}

\begin{figure}
  \centering
  \includegraphics[width=0.85\textwidth]{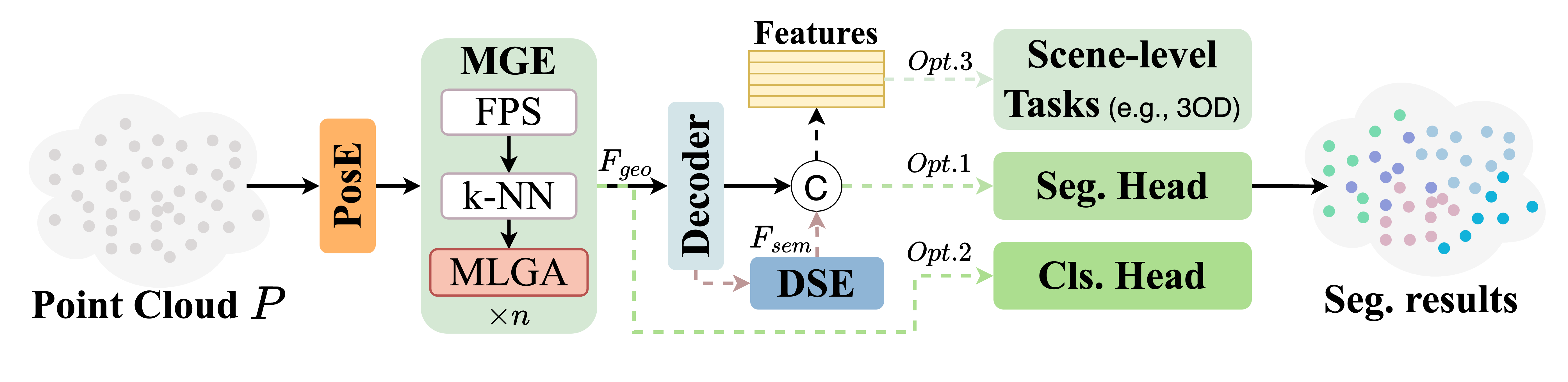} 
  \caption{\textbf{Overview of PointeNet.} It is an efficient network comprising two key modules: the multivariate geometric encoding (MGE) module and the distance-aware semantic enhancement (DSE) module. It can be flexibly combined with segmentation/classification heads to achieve excellent point cloud segmentation/classification tasks. Furthermore, it can serve as a plug-and-play module embedded in arbitrary scene-level tasks, such as 3D Object Detection (3OD), contributing to further performance enhancement.}
  \label{fig:overview}
\end{figure}

We propose an efficient network for point cloud analysis, named PointeNet. It can lightweightly capture and adaptively aggregate multivariate geometric and semantic features of point clouds using a minimal number of learnable parameters.
The detailed framework of PointeNet is illustrated in Fig. \ref{fig:overview}.

\subsection{Revisiting Point-based Methods}

Unlike grid-based methods, point-based methods have gained popularity due to their ability to directly learn the underlying point cloud representation without preprocessing. 
PointNet/PointNet++ \cite{pointnet,pointnet++} are pioneers in this paradigm, employing a hierarchical feature learning approach through the stacking of multiple learning stages.
We first delve into the core idea of PointNet++ and then explore the strengths and weaknesses of subsequent methods.

Given a set of points $\mathcal{P}=\left \{ p_{i} \mid i = 1, \dots, \mathcal{N} \right \} \in \mathbb{R}^{\mathcal{N} \times 3}$, where $\mathcal{N}$ is the number of points. 
The core idea of PointNet++ is to utilize FPS for down-sampling point clouds, use k-NN for grouping points, and then employ local geometric extractors to extract features, followed by max pooling to aggregate these features. The key steps for geometric feature extraction and aggregation can be formulated as:
\begin{equation} \label{eq1}
    f_{c} = \mathcal{A}\left ( \Phi \left ( f_{c,j} \right ) \mid j \in \mathcal{N}_{c}  \right ) 
\end{equation}
where $\mathcal{A}\left( \cdot\right)$ means aggregation function (i.e., max-pooling in PointNet++), $\Phi\left( \cdot \right)$ denotes the local feature extraction extractor (i.e., MLPs in PointNet++), and $f_{c,j}$ is the $j-$th neighbor point feature of center point $p_{c}$ within the local region. 

Regarding network architecture design, PointNet++ establishes a versatile baseline for point cloud analysis.
Later, based on this baseline, several methods are introduced, either emphasizing local geometric extractors \cite{dgcnn,3D-GCN,KPConv,paconv} or stacking repeated network blocks to deepen the network \cite{pointmlp,RSCNN}. While these methods can capture detailed local geometric information and often yield promising performance, their development encounters three challenges.

Firstly, whether by introducing intricate geometric extractors or deepening the network to enhance the representativeness of local geometric features, these methods inevitably introduce a large number of learnable parameters, leading to increased computational complexity and exacerbating the burden on CPU/GPU, resulting in inference latency.
For instance, PointMLP \cite{pointmlp} achieves 16.7 M parameters by stacking residual network blocks, and the training time is as high as 23 hours when performing segmentation tasks on ShapeNetPart.
This represents a considerable computational overhead.
Secondly, intricate local geometric extractors are maturing and even coming to saturation. 
Performance improvements on popular benchmarks like ModelNet40, ScanObjectNN, and ShapeNetPart are encountering bottlenecks.
Third, these methods are primarily designed for object-level point cloud analysis scenarios and rarely consider more realistic and valuable scene-level applications, such as autonomous driving.
These limitations prompt us to explore a new paradigm that ``lightens the load'' on local feature extractors. 
In other words, \textit{can we design an efficient lightweight point cloud analysis framework with fewer parameters that achieves satisfactory performance not only at the object level but also at the scene level?}

Excitingly, Point-NN makes the first successful attempt to perform point cloud analysis by constructing a PointNet++-style hierarchical framework using only non-learnable components such as FPS, k-NN, and pooling.
However, it relies solely on a single geometric feature (i.e., spatial neighborhood information), overlooking the more potential point cloud features, such as normals and curvatures of the local surfaces, and even semantic features. And, it also lacks customized optimization for real-world scene-level applications.

\subsection{Framework of PointeNet}

To overcome the aforementioned limitations, we propose a simple yet efficient network, named PointeNet. It can lightweightly capture and adaptively aggregate multivariate geometric and semantic features without the need for intricate or heavy operations. Moreover, it is specifically optimized for real-world autonomous driving scenarios.

As shown in Figure \ref{fig:overview}, the key operations of PointeNet can be formulated as:
\begin{equation} 
    f_{c} = \mathcal{C} \left ( f^{g}_{c}, \Phi_{sem} \left ( f^{g}_{c} \right ) \right ) 
\end{equation}
\begin{equation}
    f^{g}_{c} =  \mathcal{A}\left ( \left ( \Phi_{pos} \left ( f_{c,j} \right ), \Phi_{sur} \left ( f_{c,j} \right ) \right )  \mid j \in \mathcal{N}_{c}  \right )
\end{equation}
where $f_{c,j}$ is the $j-$th neighbor point feature of the center point $p_{c}$ within the local region. $\Phi_{pos}\left ( \cdot \right )$ and $\Phi_{sur}\left ( \cdot \right )$ are functions in the multivariate geometric encoding module responsible for extracting spatial neighboring features and curvature-normal features, respectively. 
And, $\Phi_{sem}\left ( \cdot \right )$ is a function for extracting distance-aware semantic features tailored for real-world autonomous driving scenarios.
$\mathcal{A}$ is an aggregation function, and in PointeNet, it is our proposed multivariate adaptive aggregation (MAA) module.
$\mathcal{C}$ is the concatenation operation, which concatenates the multivariate geometric features and distance-aware semantic features.

\begin{wrapfigure}{r}{0.5\textwidth}
  \centering
  \includegraphics[width=0.5\textwidth]{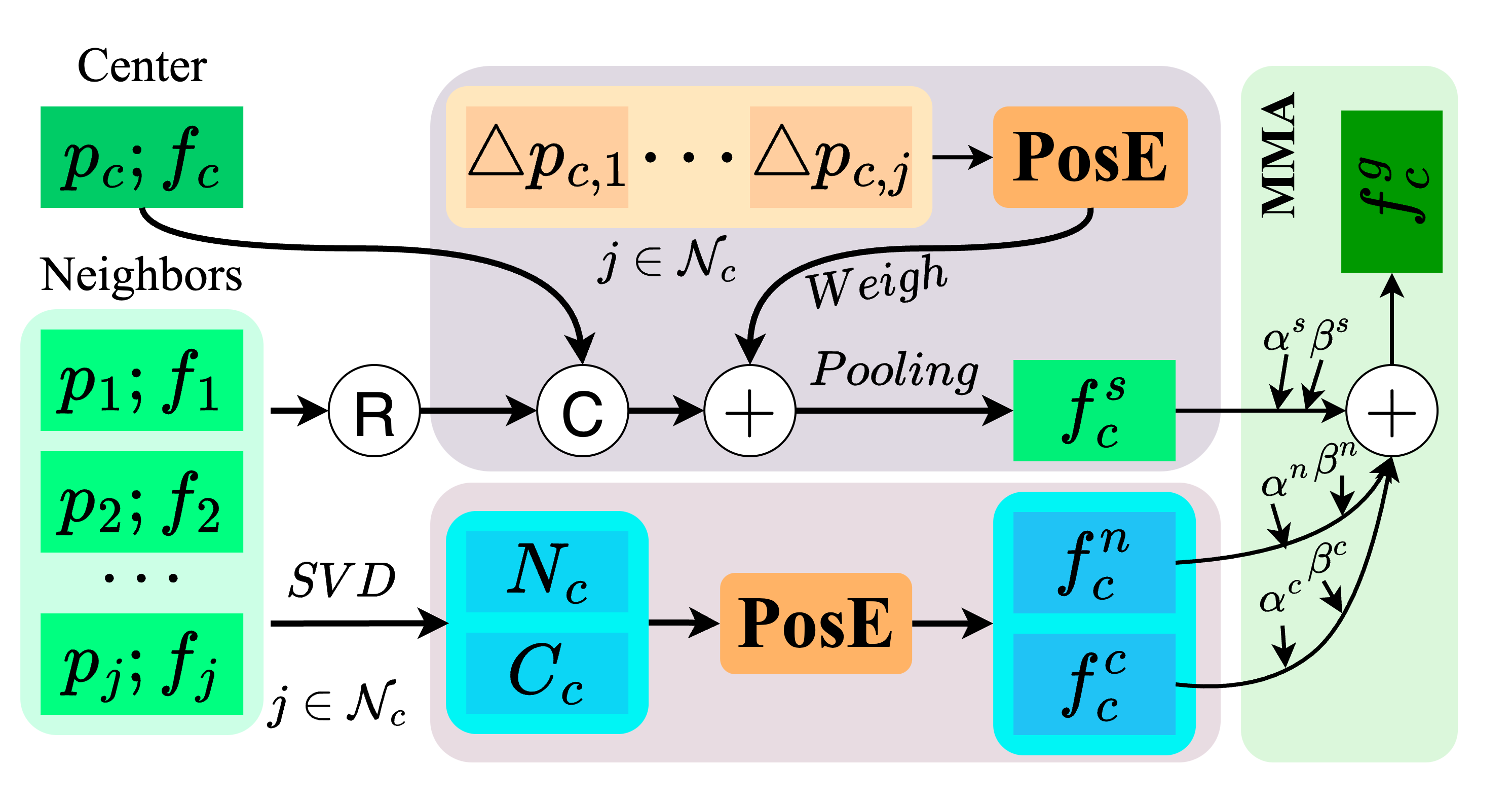} 
  \caption{\textbf{Overview of multivariate geometric aggregation (MLGA) module.} It is capable of lightweight capturing of 3D geometries, including the spatial neighboring $f^{s}_{c}$, curvature $f^{c}_{c}$, and normal $f^{n}_{c}$ features within a local region, and then undergoes a multivariate adaptive aggregation (MAA) module to adaptively Aggregate various features, finally outputting multivariate local geometric features $f^{g}_{c}$}
  \label{fig:MLGA}
\end{wrapfigure}

Our efficient PointeNet exhibits some prominent advantages:
i) PointeNet comprises only a minimal number of parameters for positional encoding, feature aggregation, and semantic segmentation. It is naturally unaffected by permutations, aligning perfectly with the characteristics of point clouds.
ii) Without the need for designing and stacking intricate feature extractors, PointeNet efficiently achieves state-of-the-art performance with only a minimal number of parameters.
iii) Additionally, due to the absence of intricate feature extractors, PointeNet is easy to train and has low hardware requirements. 

\subsection{Multivariate Geometric Encoding}

To clearly describe a given point cloud, our natural intuition leads us to seek mastery over its spatial neighboring, surface detail, and even semantic information.
However, existing methods, such as PointMLP \cite{pointmlp} and Point-NN \cite{pointnn}, almost entirely overlook the potential surface details or employ elaborate fusion strategies to aggregate various features.
In this regard, we introduce a Multivariate Geometric Encoding (MGE) module, consisting of non-parametric FPS, k-NN, and a module for multivariate local geometric aggregation (MLGA). This module is designed to efficiently capture and adaptively aggregate multivariate geometric features.

MLGA is a crucial component of MGE. As illustrated in Figure \ref{fig:MLGA}, it not only captures spatial neighboring information within the local region but also comprehensively depicts local surface details, including curvature and normal information.
Moreover, it achieves adaptive aggregation of multivariate geometric features. 
Note that a limited number of parameters in MLGA exist in the position encoding and the adaptive feature aggregation process.
In the following, we elaborate on the process of capturing and aggregating multivariate geometric features.

The encoding of spatial neighboring information in local neighborhoods, similar to Point-NN, is represented by the following formula:
\begin{equation}
    f^{s}_{c} = \left ( \mathcal{C}\left ( f_{c}, f_{j} \right )   + PosE\left ( \bigtriangleup p_{c,j} \right )  \right ) \odot PosE\left ( \bigtriangleup p_{c,j} \right ) , j \in \mathcal{N}_{c} 
\end{equation}
where $f_{c}$ and $f_{j}$ represent the center and neighbor features within the local region, respectively.
$\bigtriangleup p_{c,j}$ represents the normalized coordinates of neighboring points by the mean and standard deviation.
$PosE\left(\cdot\right)$ refers to the position encoding, and here, it is implemented using an MLPs.

Surface feature information in point clouds refers to a set of features that represent the shape of surfaces in 3D space. 
These features are crucial for point cloud analysis, and common surface feature information includes normals, curvature, smoothness, boundaries, and even texture. 
Among these, normals and curvature are the most typical surface features. GeoMAE has successfully demonstrated this by using normals and curvature of local surfaces in point clouds as self-supervised signals for pre-training, leading to improved performance in downstream tasks such as 3D detection.
Therefore, to enhance the representation of the single spatial neighboring features in the encoder, we introduce surface curvature and normal as two additional geometries. For a set of neighboring points $p_{j}$, we begin by computing the covariance matrix:
\begin{equation}
    M_{c} = \frac{1}{\mathcal{N}_{c}}\sum_{j=1}^{\mathcal{N}_{c}}p_{j}p^{T}_{j}-\bar{p}\bar{p}^{T}
\end{equation}
where $\bar{p}=\frac{1}{\mathcal{N}_{c}} \sum_{j=1}^{\mathcal{N}_{c}} p_{j}$ is the centroid of the local region. Next, we perform Singular Value Decomposition (SVD) on the covariance matrix $M_{c}$ to obtain the eigenvalues $c_{1}, c_{2}, c_{3}$ and corresponding eigenvectors $n_{1}, n_{2}, n_{3}$. Here, we use the eigenvector $n_{3}$ corresponding to the smallest eigenvalue $\lambda_{3}$ as the pseudo-normal vector $N_{c}=n_{3}$. Following that, we normalize the three eigenvalues to obtain the pseudo-curvature vector $C_{c}=\left \{ c_{0}, c_{1}, c_{2} \right \}$:
\begin{equation}
    c_{m}=\frac{c_{m}}{ {\textstyle \sum_{i=1}^{3}}c_{i} }, m \in \left \{ 1, 2, 3 \right \}
\end{equation}

As both pseudo-curvature and pseudo-normal vectors are low-dimensional, to embed them into high-dimensional geometric features, we naturally use positional encoding to map them into a higher-dimensional space, i.e., $f^{n}_{c} = PosE\left(N_{c}\right)$ and $f^{c}_{c} = PosE\left(C_{c}\right)$.

Unlike previous methods \cite{mmmot} that use MLPs with relatively more parameters to aggregate multiple features, we ingeniously introduce learnable parameters $\alpha, \beta$ to learn channel-wise weights and biases for each feature, i.e., $f^{s}_{c}$, $f^{n}_{c}$ and $f^{c}_{c}$. We then perform adaptive aggregation using the following formula:
\begin{equation}
    f^{g}_{c}=\sum_{i=\left \{ s,c,n \right \} } \left ( \alpha^{i} \odot f^{i}_{c} + \beta^{i} \right ) 
\end{equation}
where $\odot$ denotes Hadamard product and $f^{g}_{c}$ is the output of aggregated multivariate geometric features.

\subsection{Distance-ware Semantic Enhancement}

Prior point cloud analysis methods, such as PointMLP \cite{pointmlp} and Point-NN \cite{pointnn}, are previously limited to object-level classification/segmentation tasks and do not have custom optimizations for scene-level tasks, such as applications in autonomous driving scenarios.
In real-world autonomous driving applications, the field of view is more expansive, and the point cloud becomes sparser, particularly as the distance increases. This poses a challenge for effective perception, especially for distant points.

The intuitive solution is to allocate more ``attention'' to the sparser point clouds at a distance, while the relatively denser point clouds nearby require only a small amount of ``attention'' for effective perception.
PV-RCNN \cite{pv-rcnn} attempts to give more attention to foreground points through semantic segmentation. 
However, it neglects the varying segmentation difficulties in different regions of the entire point cloud scene based on their distances.
Therefore, the key is to distribute attention from the denser point clouds nearby to the sparser regions at a distance, ensuring a more equitable and easily trainable perception network.

\begin{wrapfigure}{hr}{0.3\textwidth}
  \centering
  \includegraphics[width=0.3\textwidth]{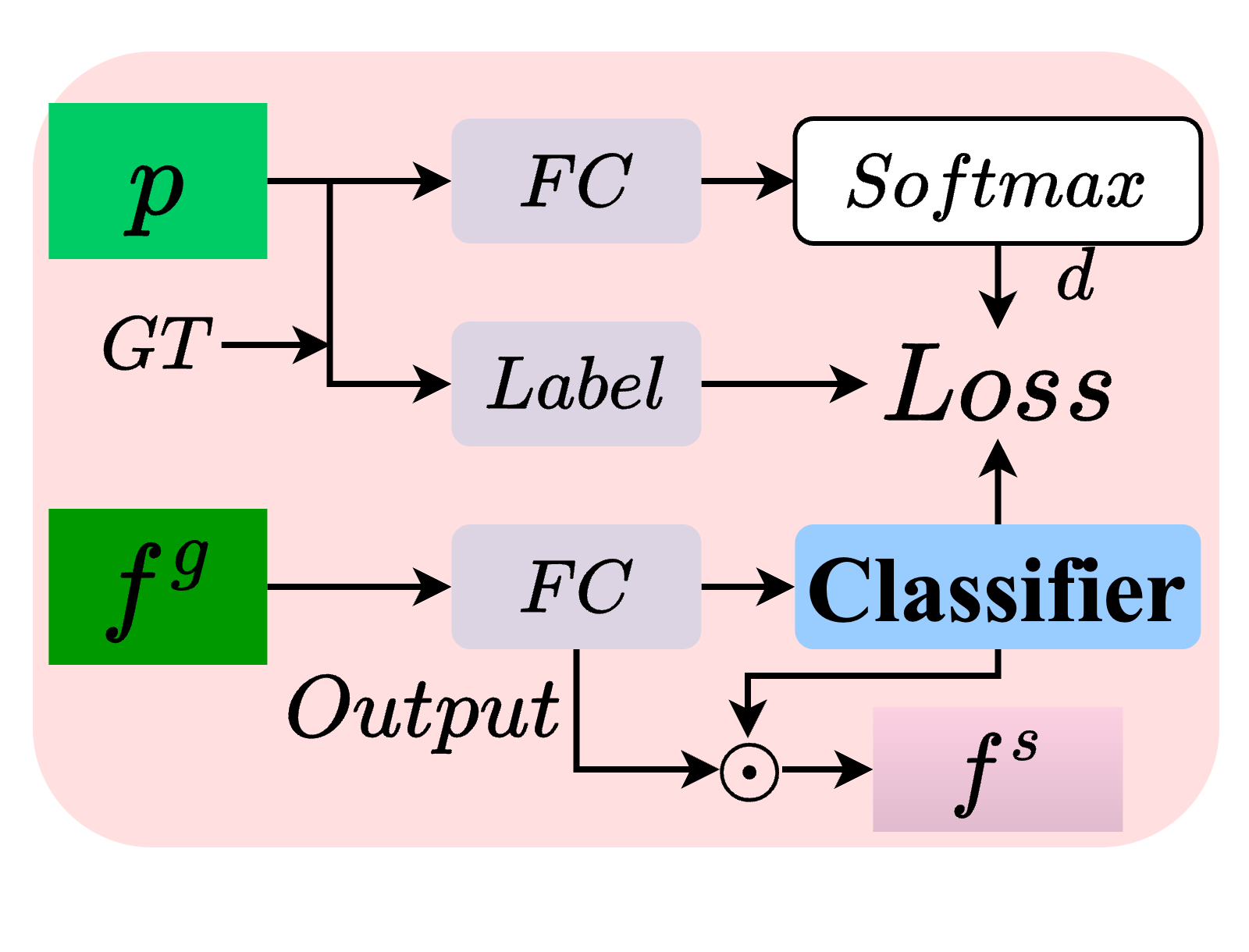} 
  \caption{\textbf{Overview of distance-ware semantic enhancement (DSE) module.} It introduces the distance factor $d$ into the semantic segmentation loss function, adaptively distinguishing the segmentation difficulty in different regions to obtain distance-aware semantic features $f^{s}_{c}$.}
  \label{fig:DSE}
\end{wrapfigure}

To address this, we introduce the Distance-aware Segmentation Enhancement (DSE) module (see Figure \ref{fig:DSE}), which incorporates the distance factor into the semantic segmentation loss, specifically using Focal Loss \cite{focalloss}. 
DSE takes the multivariate geometric features $f^{g}$ from MGE and corresponding point coordinates $p$ as input and outputs the features of the intermediate segmentation process as the semantic features $f^{s}$.
Specifically, DSE employs four fully connected (FC) layers and the \textit{Sigmoid} function to perform binary classification for each point (foreground or background). 
Additionally, it utilizes two FC layers and the \textit{Softmax} function to extract the distance factor $d \in \left[0,1\right]$ based on the absolute values of the $x, y$ coordinates for each point $p$. 
Subsequently, we incorporate $d$ into Focal Loss, replacing the original adjustment factors $\alpha$ and $\gamma$ to adaptively adjust the segmentation penalty based on the distance, as expressed by the formula:
\begin{equation}
    L_{seg}=-d\left ( 1-p_{t} \right ) ^{\frac{1}{d}}\log{p_{t}} 
\end{equation}
where $p_{t}$ denotes the probability of the predicted sample belonging to the positive class, the term $-d\left ( 1-p_{t} \right ) ^{\frac{1}{d}}$ serves as the weight for the $\log{p_{t}}$ term, with $d$ acting as an adaptive adjustment factor for this weight.
For distant points, their $x, y$ coordinates are relatively large, leading to larger distance factors $d$. Conversely, for nearby points, their distance factors $d$ are smaller. Consequently, $\frac{1}{d}$ for distant points is smaller than for nearby points, resulting in a greater penalty for distant points represented by $\left ( 1-p \right ) ^{\frac{1}{d} }\log{p_{t}}$ compared to nearby points.
Furthermore, as the distance factor $d$ increases, the penalty $\left ( 1-p \right ) ^{\frac{1}{d} }\log{p_{t}}$ also increases. 
To regulate the overall loss value, we multiply it by the distance factor $d$.

By directly concatenating multivariate geometric features and distance-aware semantic features, we utilize the combined features as auxiliary features for point clouds. This allows for a cost-effective enhancement of the performance of any 3D object detection network. 
This improvement not only requires minimal code modifications but also incurs low computational costs. \textit{Importantly, it enables the extension of point cloud analysis methods to more practically valuable scene-level tasks.}

\section{Experiments}

In this section, we conduct a comprehensive evaluation of PointeNet not only on object-level datasets such as ModelNet40 \cite{ModelNet40}, ScanObjectNN \cite{ScanObjectNN}, and ShapeNetPart \cite{ShapeNetPart}, but also validate its performance on a scene-level dataset KITTI \cite{kitti}.
Detailed ablation studies demonstrate the effectiveness and efficiency of PointeNet with both quantitative and qualitative analysis.

\subsection{Datasets}

For \textit{Part Segmentation}, we employ the ShapeNetPart \cite{ShapeNetPart} dataset, which consists of 16881 shapes with 16 classes, totaling 50 part labels. In each class, the number of parts varies from 2 to 6. 

For \textit{Shape Classification}, we evaluate the performance on the synthetic ModelNet40 \cite{ModelNet40} and the real-world ScanObjectNN \cite{ScanObjectNN} datasets.
ModelNet40 comprises 9843 training and 2468 testing meshed CAD models distributed across 40 categories.
ScanObjectNN includes 15000 objects categorized into 15 classes, featuring 2902 unique object instances in real-world scenarios.
Due to the presence of background, noise, and occlusions, ScanObjectNN poses significant challenges for existing methods.
In our experiments, we specifically consider the most challenging perturbed variant.

For \textit{3D Object Detection}, our experiments are conducted on the KITTI dataset \cite{kitti}, which is a popular benchmark for 3OD in autonomous driving scenes. It contains 7481 samples for training and 7518 samples for testing. Each sample consists of a point cloud and an RGB image with nine categories. 

\begin{figure}[t]
    \centering
    \includegraphics[width=1\linewidth]{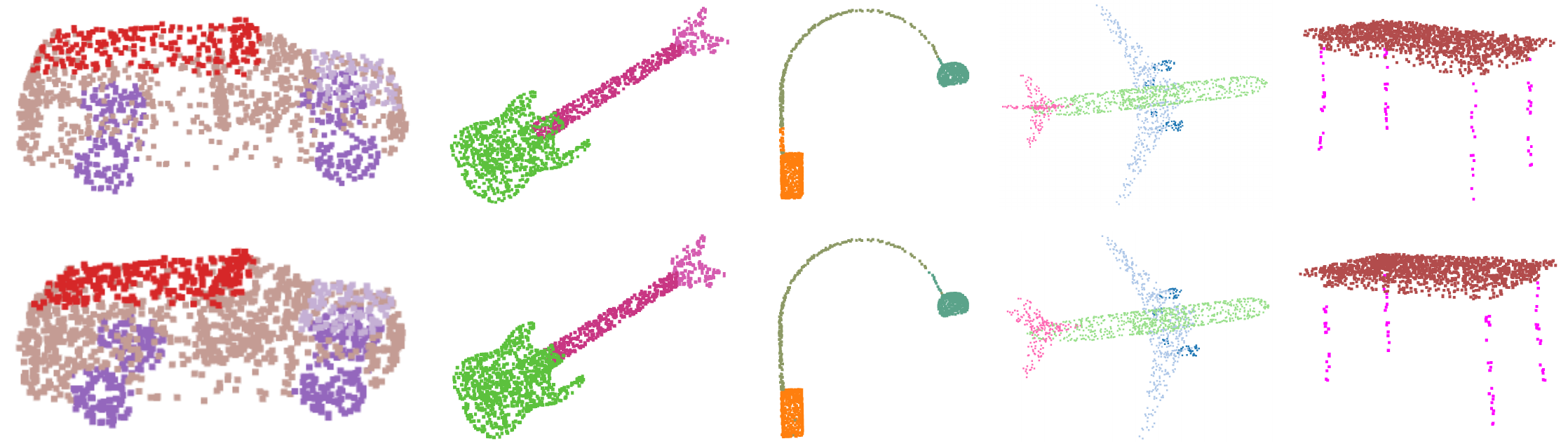}
    \caption{\textbf{Visualization of part segmentation results on ShapeNetPart}. The top line represents the ground truth, while the bottom line corresponds to our predictions. Upon visualization, our predictions closely align with the ground truth.}
    \label{fig:shapenetpart_vis}
\end{figure}

\begin{table}[h]
\caption{\textbf{Part segmentation on ShapeNetPart.} We report mean IoU scores across classes (cls. mIoU) and instances (Inst. mIoU) on the testing set. Note that the embedding dimension of PointeNet is set to 90 and $\ast$ means the reproduced results. }
\label{shapenetpart}
\centering
\scalebox{1}{
\begin{tabular}{c|cc|ccc}
\hline
Method                                                        & \begin{tabular}[c]{@{}c@{}}Cls.\\ mIoU (\%) \end{tabular} & \begin{tabular}[c]{@{}c@{}}Inst.\\ mIoU (\%) \end{tabular} & Param. & \begin{tabular}[c]{@{}c@{}}Train\\ Time\end{tabular} & \begin{tabular}[c]{@{}c@{}}Test\\ Time\end{tabular} \\ \hline
PointNet \cite{pointnet}        & 80.4            & 83.7            & 8.3 M    & -      & -             \\
PointNet++ \cite{pointnet++}    & 81.9            & 85.1            & 1.8 M    & -      & -         \\
Kd-Net \cite{Kd-Net}            & -               & 82.3            & -        & -      & -             \\
SpiderCNN \cite{SpiderCNN}      & 82.4            & 85.3            & -        & -      & -             \\
SPLATNet \cite{SPLATNet}        & 83.7            & 85.4            &  -       & -      & -             \\ \hline
PointMLP$^\ast$ \cite{pointmlp}        & 84.0            & \textbf{85.7}   & 16.7 M   & 23 h   & 7 ms          \\ 
Point-NN$^\ast$ \cite{pointnn}         & -               & 70.7            & 0 M      & 0 h    & 19 ms         \\ 
PointeNet (Ours)                 & \textbf{84.2}   & 85.6            & 8.6 M    & 8 h    & 4 ms          \\ \hline
\end{tabular}
}
\end{table}

\begin{table}[h]
\caption{\textbf{Shape classification on synthetic ModelNet40.} We report the class-average accuracy (mAcc) and overall accuracy (OA) on the testing set. $\ast$ means the reproduced results.}
\label{modelnet40}
\centering
\scalebox{1}{
\begin{tabular}{c|cc|ccc}
\hline
Method       & mAcc (\%)  & OA (\%)   & Param. & \begin{tabular}[c]{@{}c@{}}Train\\ Time\end{tabular} & \begin{tabular}[c]{@{}c@{}}Test\\ Time\end{tabular} \\ \hline
PointNet \cite{pointnet}             & 86.0           & 89.2            & -       & -      & -    \\
PointNet++ \cite{pointnet++}         & -              & 91.9            & 1.4 M   & -      & -    \\
PointCNN \cite{pointcnn}            & 88.1           & 92.5            & -       & -      & -    \\
PointConv \cite{Pointconv}           & -              & 92.5            & 18.6 M  & -      & -    \\
KPConv \cite{KPConv}                 & -              & 92.9            & 15.2 M  & -      & -    \\
Point Trans. \cite{PointTransformer} & 90.6           & 93.7            & -       & -      & -    \\
GBNet \cite{GBNet}                   & 91.0           & 93.8            & 8.4 M   & -      & -    \\ \hline
PointMLP$^\ast$ \cite{pointmlp}             & 90.8           & 93.3            & 13.2 M  & 11 h   & 5 ms  \\ 
Point-NN$^\ast$ \cite{pointnn}              & -              & 81.3            & 0 M     & 0 h    & 4 ms  \\ 
PointeNet (Ours)                      & \textbf{91.5} & \textbf{93.9}   & 1.1 M    & 4.1 h  & 3 ms  \\ \hline
\end{tabular}
}
\end{table}

\subsection{Part Segmentation on ShapeNetPart}
We present the results of PointeNet and compare our method with several recent works, including PointNet \cite{pointnet}, PointNet++ \cite{pointnet++}, Kd-Net \cite{Kd-Net}, SpiderCNN \cite{SpiderCNN}, SPLATNet \cite{SPLATNet}, PointMLP \cite{pointmlp} and Point-NN \cite{pointnn}, for the 3D shape part segmentation task on the ShapeNetPart benchmark, as summarized in Table \ref{shapenetpart}.

Analyzing the results, our PointeNet achieves segmentation performance comparable to PointMLP while utilizing only half of its parameters (8.6 M vs. 16.7 M), especially surpassing PointMLP by 0.2\% in cls. mIoU (84.2 \% vs. 84.0 \%). 
Additionally, our PointeNet exhibits a clear advantage over PointMLP in terms of both training and testing times. Specifically, the training time is only one-third of PointMLP (8.6 h vs. 23 h), and the testing time is also twice as fast as PointMLP (4 ms vs. 7 ms).
Compared to the non-parametric Point-NN, our method demonstrates stronger segmentation capabilities, with an Inst. mIoU metric surpassing 14.9\%, and the testing time is only one-fifth of Point-NN (4 ms vs. 19 ms).

Furthermore, we also present visualizations of the ground truth and predicted segmentation in Figure \ref{fig:shapenetpart_vis}. Intuitively, the predictions from our PointeNet closely match the ground truth, providing qualitative evidence for the validity of our method.

\begin{table}[h]
\caption{\textbf{Shape classification on the real-world ScanObjectNN.} We report the class-average accuracy (mAcc) and overall accuracy (OA) on the most challenging variant of the testing set. $\ast$ means the reproduced results.}
\label{ScanObjectNN}
\centering
\scalebox{1}{
\begin{tabular}{c|cc|ccc}
\hline
Method     & mAcc (\%) & OA (\%)   & Param. & \begin{tabular}[c]{@{}c@{}}Train\\ Time\end{tabular} & \begin{tabular}[c]{@{}c@{}}Test\\ Time\end{tabular} \\ \hline
PointNet \cite{pointnet}     & 63.4           & 68.2           & 3.5 M  & -       & -       \\
SpiderCNN \cite{SpiderCNN}   & 69.8           & 73.7           & -      & -       & -       \\
PointNet++ \cite{pointnet++} & 75.4           & 77.9           & 1.7 M  & -       & -       \\
PointCNN \cite{pointcnn}     & 75.1           & 78.5           & 15.2 M & -       & -       \\
GBNet \cite{GBNet}           & 77.8           & 80.5           & 8.4 M  & -       & -       \\
SimpleView \cite{SimpleView} & -              & 80.5           & -      & -       & -       \\\hline
PointMLP$^\ast$ \cite{pointmlp}     & 83.2           & 85.0           & 13.2 M & 7.6 h   & 4 ms   \\ 
Point-NN$^\ast$ \cite{pointnn}      & -              & 64.8           & 0 M    & 0 h     & 4 ms       \\
PointeNet (Ours)              & \textbf{86.4}  & \textbf{87.9}  & 4.7 M  & 12.3 h  & 5 ms   \\ \hline
\end{tabular}
}
\end{table}

\subsection{Shape Classification on ModelNet40}

We also evaluate PointeNet and compare our method with several recent works, including PointNet \cite{pointnet}, PointNet++ \cite{pointnet++}, PointCNN \cite{pointcnn}, PointConv \cite{Pointconv}, KPConv \cite{KPConv}, PointTransformer \cite{PointTransformer}, GBNet \cite{GBNet}, PointMLP \cite{pointmlp} and Point-NN \cite{pointnn}, for the 3D shape classification task on the synthetic ModelNet40 \cite{ModelNet40} benchmark.
Following the standard practice in the community, we report the class-average accuracy (mAcc) and overall accuracy (OA) on the testing set in Table \ref{modelnet40}. 

Our PointeNet outperforms PointMLP by 0.6\% in OA (93.9\% vs. 93.3\%) and 0.7\% in mAcc (91.5\% vs. 90.8\%), all while requiring 12 times fewer parameters (13.2 M vs. 1.1 M). Moreover, in terms of training and testing times, our method is more cost-effective, with half the training time (4.1 h vs. 11 h) and approximately twice the testing speed (3 ms vs. 5 ms) compared to PointMLP.
Compared to Point-NN, our method not only demonstrates superior performance in OA (93.9\% vs. 81.3\%) but also achieves faster testing speeds (3 ms vs. 4 ms).
These experimental results validate the efficiency and superiority of our method in shape classification tasks.

\subsection{Shape Classification on ScanObjectNN}

Although our method has already demonstrated its efficiency in shape classification tasks on the synthetic ModelNet40 dataset, to thoroughly validate this capability, we extend our evaluation to the ScanObjectNN benchmark \cite{ScanObjectNN}, where we compare PointeNet with several recent methods, including PointNet \cite{pointnet}, PointNet++ \cite{pointnet++}, PointCNN \cite{pointcnn}, SpiderCNN \cite{SpiderCNN}, GBNet \cite{GBNet}, SimpleView \cite{SimpleView}, PointMLP \cite{pointmlp} and Point-NN \cite{pointnn}.

As shown in Table \ref{ScanObjectNN}, our PointMLP surpasses all methods with a significant improvement in both class mean accuracy (mAcc) and overall accuracy (OA). 
Notably, despite utilizing only one-third of the parameters (4.7 M vs. 13.2 M) employed by PointMLP, our method achieves significantly superior performance. Specifically, we surpass PointMLP by more than 3.2\% in mAcc (86.4\% vs. 83.2\%) and 2.9\% in OA (87.9\% vs. 85.0\%).
Compared to Point-NN, our method also maintains a significant advantage.
In addition, it's worth noting that our method achieves a smaller gap between class average accuracy (mAcc) and overall accuracy (OA) compared to PointMLP (1.5\% vs. 1.8\%). This observation suggests that PointeNet avoids bias towards any specific category, demonstrating commendable robustness.
Through these results on the real-world ScanObjectNN, the efficiency and superiority of our method in the shape classification task have been thoroughly validated.

\begin{table}[ht]
\caption{Results of PointRCNN with and without PointeNet on the KITTI val set. The best performance value is in \textbf{bold}, second-best is \underline{underlined}. $\ast$ means the reproduced results.}
\label{kittival}
\centering
\scalebox{0.9}{
\begin{tabular}{c|ccc|ccc|ccc}
\hline
\multirow{2}{*}{Method} & \multicolumn{3}{c|}{Car (3D AP$_{R40}$)} & \multicolumn{3}{c|}{Pedestrian (3D AP$_{R40})$} & \multicolumn{3}{c}{Cyclist (3D AP$_{R40}$)} \\
                        & Easy   & Mod.   & Hard   & Easy      & Mod.     & Hard     & Easy    & Mod.    & Hard    \\ \hline
SECOND \cite{second}                                                        & 90.55    & 81.61    & 78.56    & 55.94       & 51.15      & 46.17      & 82.97     & 66.74     & 62.78     \\
PointPillars \cite{pointpillars}                                                  & 87.75    & 78.41    & 75.19    & 57.30       & 51.42      & 46.87      & 81.57     & 62.93     & 58.98     \\
PV-RCNN   \cite{pv-rcnn}                                                  & \underline{92.10}    & \textbf{84.36}    & \textbf{82.48}    & \underline{64.26}       & \underline{56.67}      & \underline{51.91}      & 88.88     & 71.95     & 66.78     \\
PC-RGNN  \cite{pc-rgnn}                                                     & 90.94    & 81.43    & 80.45    & -           & -          & -          & -         & -         & -         \\
Voxel-RCNN  \cite{voxel-rcnn}                                                  & 91.72    & \underline{83.19}    & 78.60    & -           & -          & -          & -         & -         & -         \\ \hline
PointRCNN$^\ast$  \cite{PointRCNN}                                                   & 91.28    & 80.78    & 78.37    & 65.12       & 56.43      & 49.24      & \underline{90.02}     & \textbf{72.42}     & \underline{67.44}     \\
PointRCNN + PointeNet & \textbf{92.34}    & 83.13    & \underline{80.78}    & \textbf{66.00}       & \textbf{58.62}      & \textbf{52.09}      & \textbf{92.99}     & \textbf{73.97}     & \textbf{69.50}     \\
$Improvement$                                                      & \textbf{1.06} & \textbf{2.35} & \textbf{2.41} & \textbf{0.88} & \textbf{2.19} & \textbf{2.85} & \textbf{2.97} & \textbf{1.55} & \textbf{2.06} \\ \hline
\end{tabular}
}
\end{table}

\begin{table}[ht]
\caption{Ablation results of the proposed MLGA on the ShapeNetPart testing set. We analyze the effect of the multivariate geometries, including spatial neighboring, normal, and curvature, as well as the multivariate feature aggregation strategy (i.e., MAA or concatenation), on PointeNet. Note that the embedding dimension of PointeNet is set to 36.}
\label{MLGA}
\centering
\scalebox{1}{
\begin{tabular}{ccc|cc|cc|ccc}
\hline
\multicolumn{3}{c|}{Geometries} & \multicolumn{2}{c|}{Aggregation} & \multirow{2}{*}{\begin{tabular}[c]{@{}c@{}}Cls.\\ mIoU (\%)\end{tabular}} & \multirow{2}{*}{\begin{tabular}[c]{@{}c@{}}Inst.\\ mIoU (\%)\end{tabular}} & \multirow{2}{*}{Param.} & \multirow{2}{*}{\begin{tabular}[c]{@{}c@{}}Train\\ Time\end{tabular}} & \multirow{2}{*}{\begin{tabular}[c]{@{}c@{}}Test\\ Time\end{tabular}} \\
Distrib.  & Normal  & Curvature & MAA           & Concat.          &                                                                      &                                                                       &                         &                                                                       &                                                                      \\ \hline
\checkmark     &         &           & \checkmark             &                  & 82.98     & 94.76    & 4.6 M    & 5.8 h     & 3 ms               \\
\checkmark         & \checkmark       &           & \checkmark             &         & 83.47       & 84.94           & 5.5 M             & 6.1 h              & 3 ms                     \\
\checkmark         &         & \checkmark         & \checkmark             &                  & 83.48          & 85.13      & 5.5 M                   & 6.1 h      & 3 ms                     \\
\checkmark         & \checkmark       & \checkmark         &               & \checkmark                & 83.65    & 85.09     & 4.6 M    & 6.1 h      & 3 ms     \\
\checkmark         & \checkmark       & \checkmark         & \checkmark             &                  & \textbf{83.73}         & \textbf{85.17}       & 5.5 M             & 6.2 h         & 3 ms               \\ \hline
\end{tabular}
}
\end{table}

\subsubsection{3D Oject Detection on KITTI}

To validate the practicality and efficiency of our method in real-world, scene-level applications, we utilize the classic PointRCNN as a baseline and evaluate it on the real-world autonomous driving dataset KITTI \cite{kitti}. We compare PointRCNN with several established 3D object detection networks, including SECOND \cite{second}, PointPillars \cite{pointpillars}, PV-RCNN \cite{pv-rcnn}, PC-RGNN \cite{pc-rgnn}, and Voxel-RCNN \cite{voxel-rcnn}. 

From the results of Table \ref{kittival}, we can observe that our PointFPN significantly improves the performance of PointRCNN in all three categories: \textit{Car}, \textit{Pedestrian}, and \textit{Cyclist}, with the highest improvement approaching 3\%. 
Moreover, PointRCNN with the addition of PointeNet also has superior performance compared to the PV-RCNN, which combines the advantageous features of Voxel and Point.
From these results, it is evident that our method performs exceptionally well in real-world application scenarios and exhibits high efficiency.

\subsection{Ablation Study}


\subsubsection{Effects of MLGA}

Our multivariate geometric encoding (MGE) module consists of non-parametric FPS, k-NN, and a multivariate local geometric aggregation (MLGA) module with a minimal number of learnable parameters, where MLGA is crucial. The proposed MLGA captures multivariate geometric features, including spatial neighboring, curvature, and normal information, and adaptively aggregates these geometric features. 
To thoroughly validate the effectiveness of these geometric features and the proposed multivariate adaptive aggregation (MAA) module, we conduct comprehensive experiments, and the results are presented in Table \ref{MLGA}.
From the results in the first three rows of Table \ref{fig:MLGA}, it can be observed that spatial geometry, curvature, and normal information, these three geometric features, all make significant contributions to PointeNet.
The comparative results in the fourth and fifth rows also demonstrate that these geometries, when adaptively aggregated through our MAA module, are more effective compared to directly concatenating them together.

\begin{table}[t]
\caption{Ablation results of the proposed DSE on the KITTI val set. We analyze the effect of MGE and DSE (with/without the distance factor $d$) on PointeNet.}
\label{DSE}
\centering
\scalebox{1}{
\begin{tabular}{c|c|cc|ccc}
\hline
\multirow{2}{*}{PointRCNN} & \multirow{2}{*}{MGE} & \multicolumn{2}{c|}{DSE} & \multicolumn{3}{c}{Car (3D AP$_{R40}$)} \\
                           &                      & w/ $d$       & w/o $d$       & Easy     & Mod.     & Hard    \\ \hline
\checkmark                 &                      &            &             & 91.28    & 80.78    & 78.37   \\
\checkmark                 & \checkmark           &            &             & 91.98    & 82.35    & 80.39   \\
\checkmark                 & \checkmark           & \checkmark &             & 92.04    & 82.51    & 80.49   \\
\checkmark                 & \checkmark           & \checkmark & \checkmark  & \textbf{92.34}    & \textbf{83.13}    & \textbf{80.78}   \\ \hline
\end{tabular}
}
\end{table}

\subsubsection{Effects of DSE}

We conduct ablation experiments on the challenging \textit{Car} category of the KITTI dataset to analyze the effectiveness of MGE and DSE (with/without the distance factor $d$) in PointeNet.
From the results of Table \ref{DSE}, the multivariate geometric features output by MGE significantly improve the performance of PointRCNN, especially with a notable enhancement of 2.02\% on the easy level.
Without the distance factor to adaptively adjust the segmentation difficulty for semantic features, there is a slight improvement of 0.1\% across the three difficulty levels.
When DSE is improved with the distance factor, PointeNet assists PointRCNN in further improvement, particularly enhancing performance by 0.62\% on the moderate level.


\section{Conclusion}

While prior methods in point cloud analysis have demonstrated promising success, they often face challenges associated with increased complexity, limiting their applicability to real-world scene-level scenarios and posing efficiency concerns. In response to these challenges, this paper introduces a lightweight network specifically optimized for scalability in real-world autonomous driving applications, offering a more efficient approach to point cloud analysis.
Our proposed network leverages non-learnable components, such as FPS and K-NN, and introduces a minimal number of learnable parameters through the proposed Multivariate Geometric Encoding (MGE) module. This module adeptly captures and adaptively aggregates multivariate geometric features, ensuring a lightweight yet comprehensive representation of the point cloud.
Furthermore, we incorporate a Distance-Aware Semantic Enhancement (DSE) module tailored for autonomous driving scenarios, capturing distance-aware semantic features. 
The integration of both multivariate geometric and semantic features contributes to a more efficient and superior performance in point cloud analysis.
In future work, we aim to explore more potential point cloud features to further enhance the capabilities of our lightweight network.

\bibliography{main}

\begin{thebibliography}{39}
\expandafter\ifx\csname natexlab\endcsname\relax\def\natexlab#1{#1}\fi
\providecommand{\url}[1]{\texttt{#1}}
\providecommand{\href}[2]{#2}
\providecommand{\path}[1]{#1}
\providecommand{\DOIprefix}{doi:}
\providecommand{\ArXivprefix}{arXiv:}
\providecommand{\URLprefix}{URL: }
\providecommand{\Pubmedprefix}{pmid:}
\providecommand{\doi}[1]{\href{http://dx.doi.org/#1}{\path{#1}}}
\providecommand{\Pubmed}[1]{\href{pmid:#1}{\path{#1}}}
\providecommand{\bibinfo}[2]{#2}
\ifx\xfnm\relax \def\xfnm[#1]{\unskip,\space#1}\fi
\bibitem[{Deng et~al.(2021)Deng, Shi, Li, Zhou, Zhang and Li}]{voxel-rcnn}
\bibinfo{author}{Deng, J.}, \bibinfo{author}{Shi, S.}, \bibinfo{author}{Li, P.}, \bibinfo{author}{Zhou, W.}, \bibinfo{author}{Zhang, Y.}, \bibinfo{author}{Li, H.}, \bibinfo{year}{2021}.
\newblock \bibinfo{title}{Voxel r-cnn: Towards high performance voxel-based 3d object detection}, in: \bibinfo{booktitle}{Proceedings of the AAAI Conference on Artificial Intelligence}, pp. \bibinfo{pages}{1201--1209}.
\bibitem[{Dosovitskiy et~al.(2021)Dosovitskiy, Beyer, Kolesnikov, Weissenborn, Zhai, Unterthiner, Dehghani, Minderer, Heigold, Gelly, Uszkoreit and Houlsby}]{dosovitskiy2020image}
\bibinfo{author}{Dosovitskiy, A.}, \bibinfo{author}{Beyer, L.}, \bibinfo{author}{Kolesnikov, A.}, \bibinfo{author}{Weissenborn, D.}, \bibinfo{author}{Zhai, X.}, \bibinfo{author}{Unterthiner, T.}, \bibinfo{author}{Dehghani, M.}, \bibinfo{author}{Minderer, M.}, \bibinfo{author}{Heigold, G.}, \bibinfo{author}{Gelly, S.}, \bibinfo{author}{Uszkoreit, J.}, \bibinfo{author}{Houlsby, N.}, \bibinfo{year}{2021}.
\newblock \bibinfo{title}{An image is worth 16x16 words: Transformers for image recognition at scale}, in: \bibinfo{booktitle}{9th International Conference on Learning Representations}.
\bibitem[{Felzenszwalb and Huttenlocher(2004)}]{felzenszwalb2004efficient}
\bibinfo{author}{Felzenszwalb, P.F.}, \bibinfo{author}{Huttenlocher, D.P.}, \bibinfo{year}{2004}.
\newblock \bibinfo{title}{Efficient graph-based image segmentation}.
\newblock \bibinfo{journal}{Int. J. Comput. Vis.} \bibinfo{volume}{59}, \bibinfo{pages}{167--181}.
\bibitem[{Geiger et~al.(2012)Geiger, Lenz and Urtasun}]{kitti}
\bibinfo{author}{Geiger, A.}, \bibinfo{author}{Lenz, P.}, \bibinfo{author}{Urtasun, R.}, \bibinfo{year}{2012}.
\newblock \bibinfo{title}{Are we ready for autonomous driving? the {KITTI} vision benchmark suite}, in: \bibinfo{booktitle}{{IEEE} Conference on Computer Vision and Pattern Recognition}, pp. \bibinfo{pages}{3354--3361}.
\bibitem[{Goyal et~al.(2021)Goyal, Law, Liu, Newell and Deng}]{SimpleView}
\bibinfo{author}{Goyal, A.}, \bibinfo{author}{Law, H.}, \bibinfo{author}{Liu, B.}, \bibinfo{author}{Newell, A.}, \bibinfo{author}{Deng, J.}, \bibinfo{year}{2021}.
\newblock \bibinfo{title}{Revisiting point cloud shape classification with a simple and effective baseline}, in: \bibinfo{booktitle}{Proceedings of the 38th International Conference on Machine Learning}, pp. \bibinfo{pages}{3809--3820}.
\bibitem[{Guo et~al.(2021)Guo, Cai, Liu, Mu, Martin and Hu}]{PCT}
\bibinfo{author}{Guo, M.}, \bibinfo{author}{Cai, J.}, \bibinfo{author}{Liu, Z.}, \bibinfo{author}{Mu, T.}, \bibinfo{author}{Martin, R.R.}, \bibinfo{author}{Hu, S.}, \bibinfo{year}{2021}.
\newblock \bibinfo{title}{{PCT:} point cloud transformer}.
\newblock \bibinfo{journal}{Comput. Vis. Media} \bibinfo{volume}{7}, \bibinfo{pages}{187--199}.
\bibitem[{He et~al.(2016)He, Zhang, Ren and Sun}]{resnet}
\bibinfo{author}{He, K.}, \bibinfo{author}{Zhang, X.}, \bibinfo{author}{Ren, S.}, \bibinfo{author}{Sun, J.}, \bibinfo{year}{2016}.
\newblock \bibinfo{title}{Deep residual learning for image recognition}, in: \bibinfo{booktitle}{{IEEE} Conference on Computer Vision and Pattern Recognition}, pp. \bibinfo{pages}{770--778}.
\bibitem[{Klokov and Lempitsky(2017)}]{Kd-Net}
\bibinfo{author}{Klokov, R.}, \bibinfo{author}{Lempitsky, V.S.}, \bibinfo{year}{2017}.
\newblock \bibinfo{title}{Escape from cells: Deep kd-networks for the recognition of 3d point cloud models}, in: \bibinfo{booktitle}{{IEEE} International Conference on Computer Vision}, pp. \bibinfo{pages}{863--872}.
\bibitem[{Lang et~al.(2019)Lang, Vora, Caesar, Zhou, Yang and Beijbom}]{pointpillars}
\bibinfo{author}{Lang, A.H.}, \bibinfo{author}{Vora, S.}, \bibinfo{author}{Caesar, H.}, \bibinfo{author}{Zhou, L.}, \bibinfo{author}{Yang, J.}, \bibinfo{author}{Beijbom, O.}, \bibinfo{year}{2019}.
\newblock \bibinfo{title}{Pointpillars: Fast encoders for object detection from point clouds}, in: \bibinfo{booktitle}{{IEEE} Conference on Computer Vision and Pattern Recognition}, pp. \bibinfo{pages}{12697--12705}.
\bibitem[{Li et~al.(2018)Li, Bu, Sun, Wu, Di and Chen}]{pointcnn}
\bibinfo{author}{Li, Y.}, \bibinfo{author}{Bu, R.}, \bibinfo{author}{Sun, M.}, \bibinfo{author}{Wu, W.}, \bibinfo{author}{Di, X.}, \bibinfo{author}{Chen, B.}, \bibinfo{year}{2018}.
\newblock \bibinfo{title}{Pointcnn: Convolution on x-transformed points}, in: \bibinfo{booktitle}{Advances in Neural Information Processing Systems 31}, pp. \bibinfo{pages}{828--838}.
\bibitem[{Liang et~al.(2022)Liang, Xie, Yu, Xia, Lin, Wang, Tang, Wang and Tang}]{bevfusion}
\bibinfo{author}{Liang, T.}, \bibinfo{author}{Xie, H.}, \bibinfo{author}{Yu, K.}, \bibinfo{author}{Xia, Z.}, \bibinfo{author}{Lin, Z.}, \bibinfo{author}{Wang, Y.}, \bibinfo{author}{Tang, T.}, \bibinfo{author}{Wang, B.}, \bibinfo{author}{Tang, Z.}, \bibinfo{year}{2022}.
\newblock \bibinfo{title}{Bevfusion: {A} simple and robust lidar-camera fusion framework}, in: \bibinfo{booktitle}{Advances in Neural Information Processing Systems}, pp. \bibinfo{pages}{10421--10434}.
\bibitem[{Lin et~al.(2020)Lin, Goyal, Girshick, He and Doll{\'{a}}r}]{focalloss}
\bibinfo{author}{Lin, T.}, \bibinfo{author}{Goyal, P.}, \bibinfo{author}{Girshick, R.B.}, \bibinfo{author}{He, K.}, \bibinfo{author}{Doll{\'{a}}r, P.}, \bibinfo{year}{2020}.
\newblock \bibinfo{title}{Focal loss for dense object detection}.
\newblock \bibinfo{journal}{{IEEE} Trans. Pattern Anal. Mach. Intell.} \bibinfo{volume}{42}, \bibinfo{pages}{318--327}.
\bibitem[{Lin et~al.(2022)Lin, Huang and Wang}]{3D-GCN}
\bibinfo{author}{Lin, Z.}, \bibinfo{author}{Huang, S.}, \bibinfo{author}{Wang, Y.F.}, \bibinfo{year}{2022}.
\newblock \bibinfo{title}{Learning of 3d graph convolution networks for point cloud analysis}.
\newblock \bibinfo{journal}{{IEEE} Trans. Pattern Anal. Mach. Intell.} \bibinfo{volume}{44}, \bibinfo{pages}{4212--4224}.
\bibitem[{Liu et~al.(2019)Liu, Fan, Xiang and Pan}]{RSCNN}
\bibinfo{author}{Liu, Y.}, \bibinfo{author}{Fan, B.}, \bibinfo{author}{Xiang, S.}, \bibinfo{author}{Pan, C.}, \bibinfo{year}{2019}.
\newblock \bibinfo{title}{Relation-shape convolutional neural network for point cloud analysis}, in: \bibinfo{booktitle}{{IEEE} Conference on Computer Vision and Pattern Recognition}, pp. \bibinfo{pages}{8895--8904}.
\bibitem[{Ma et~al.(2022)Ma, Qin, You, Ran and Fu}]{pointmlp}
\bibinfo{author}{Ma, X.}, \bibinfo{author}{Qin, C.}, \bibinfo{author}{You, H.}, \bibinfo{author}{Ran, H.}, \bibinfo{author}{Fu, Y.}, \bibinfo{year}{2022}.
\newblock \bibinfo{title}{Rethinking network design and local geometry in point cloud: {A} simple residual {MLP} framework}, in: \bibinfo{booktitle}{The Tenth International Conference on Learning Representations}.
\bibitem[{Qi et~al.(2019)Qi, Litany, He and Guibas}]{votenet}
\bibinfo{author}{Qi, C.R.}, \bibinfo{author}{Litany, O.}, \bibinfo{author}{He, K.}, \bibinfo{author}{Guibas, L.J.}, \bibinfo{year}{2019}.
\newblock \bibinfo{title}{Deep hough voting for 3d object detection in point clouds}, in: \bibinfo{booktitle}{{IEEE} International Conference on Computer Vision}, pp. \bibinfo{pages}{9276--9285}.
\bibitem[{Qi et~al.(2017a)Qi, Su, Mo and Guibas}]{pointnet}
\bibinfo{author}{Qi, C.R.}, \bibinfo{author}{Su, H.}, \bibinfo{author}{Mo, K.}, \bibinfo{author}{Guibas, L.J.}, \bibinfo{year}{2017}a.
\newblock \bibinfo{title}{Pointnet: Deep learning on point sets for 3d classification and segmentation}, in: \bibinfo{booktitle}{{IEEE} Conference on Computer Vision and Pattern Recognition}, pp. \bibinfo{pages}{77--85}.
\bibitem[{Qi et~al.(2017b)Qi, Yi, Su and Guibas}]{pointnet++}
\bibinfo{author}{Qi, C.R.}, \bibinfo{author}{Yi, L.}, \bibinfo{author}{Su, H.}, \bibinfo{author}{Guibas, L.J.}, \bibinfo{year}{2017}b.
\newblock \bibinfo{title}{Pointnet++: Deep hierarchical feature learning on point sets in a metric space}, in: \bibinfo{booktitle}{Advances in Neural Information Processing Systems}, pp. \bibinfo{pages}{5099--5108}.
\bibitem[{Qiu et~al.(2022)Qiu, Anwar and Barnes}]{GBNet}
\bibinfo{author}{Qiu, S.}, \bibinfo{author}{Anwar, S.}, \bibinfo{author}{Barnes, N.}, \bibinfo{year}{2022}.
\newblock \bibinfo{title}{Geometric back-projection network for point cloud classification}.
\newblock \bibinfo{journal}{{IEEE} Trans. Multim.} \bibinfo{volume}{24}, \bibinfo{pages}{1943--1955}.
\bibitem[{Ran et~al.(2021)Ran, Zhuo, Liu and Lu}]{ran2021learning}
\bibinfo{author}{Ran, H.}, \bibinfo{author}{Zhuo, W.}, \bibinfo{author}{Liu, J.}, \bibinfo{author}{Lu, L.}, \bibinfo{year}{2021}.
\newblock \bibinfo{title}{Learning inner-group relations on point clouds}, in: \bibinfo{booktitle}{2021 {IEEE/CVF} International Conference on Computer Vision}, pp. \bibinfo{pages}{15457--15467}.
\bibitem[{Shi et~al.(2020)Shi, Guo, Jiang, Wang, Shi, Wang and Li}]{pv-rcnn}
\bibinfo{author}{Shi, S.}, \bibinfo{author}{Guo, C.}, \bibinfo{author}{Jiang, L.}, \bibinfo{author}{Wang, Z.}, \bibinfo{author}{Shi, J.}, \bibinfo{author}{Wang, X.}, \bibinfo{author}{Li, H.}, \bibinfo{year}{2020}.
\newblock \bibinfo{title}{{PV-RCNN:} point-voxel feature set abstraction for 3d object detection}, in: \bibinfo{booktitle}{2020 {IEEE/CVF} Conference on Computer Vision and Pattern Recognition, {CVPR} 2020, Seattle, WA, USA, June 13-19, 2020}, pp. \bibinfo{pages}{10526--10535}.
\bibitem[{Shi et~al.(2019)Shi, Wang and Li}]{PointRCNN}
\bibinfo{author}{Shi, S.}, \bibinfo{author}{Wang, X.}, \bibinfo{author}{Li, H.}, \bibinfo{year}{2019}.
\newblock \bibinfo{title}{Pointrcnn: 3d object proposal generation and detection from point cloud}, in: \bibinfo{booktitle}{{IEEE} Conference on Computer Vision and Pattern Recognition}, pp. \bibinfo{pages}{770--779}.
\bibitem[{Su et~al.(2018)Su, Jampani, Sun, Maji, Kalogerakis, Yang and Kautz}]{SPLATNet}
\bibinfo{author}{Su, H.}, \bibinfo{author}{Jampani, V.}, \bibinfo{author}{Sun, D.}, \bibinfo{author}{Maji, S.}, \bibinfo{author}{Kalogerakis, E.}, \bibinfo{author}{Yang, M.}, \bibinfo{author}{Kautz, J.}, \bibinfo{year}{2018}.
\newblock \bibinfo{title}{Splatnet: Sparse lattice networks for point cloud processing}, in: \bibinfo{booktitle}{2018 {IEEE} Conference on Computer Vision and Pattern Recognition}, pp. \bibinfo{pages}{2530--2539}.
\bibitem[{Thomas et~al.(2019)Thomas, Qi, Deschaud, Marcotegui, Goulette and Guibas}]{KPConv}
\bibinfo{author}{Thomas, H.}, \bibinfo{author}{Qi, C.R.}, \bibinfo{author}{Deschaud, J.}, \bibinfo{author}{Marcotegui, B.}, \bibinfo{author}{Goulette, F.}, \bibinfo{author}{Guibas, L.J.}, \bibinfo{year}{2019}.
\newblock \bibinfo{title}{Kpconv: Flexible and deformable convolution for point clouds}, in: \bibinfo{booktitle}{2019 {IEEE/CVF} International Conference on Computer Vision}, pp. \bibinfo{pages}{6410--6419}.
\bibitem[{Uy et~al.(2019)Uy, Pham, Hua, Nguyen and Yeung}]{ScanObjectNN}
\bibinfo{author}{Uy, M.A.}, \bibinfo{author}{Pham, Q.}, \bibinfo{author}{Hua, B.}, \bibinfo{author}{Nguyen, D.T.}, \bibinfo{author}{Yeung, S.}, \bibinfo{year}{2019}.
\newblock \bibinfo{title}{Revisiting point cloud classification: {A} new benchmark dataset and classification model on real-world data}, in: \bibinfo{booktitle}{2019 {IEEE/CVF} International Conference on Computer Vision}, pp. \bibinfo{pages}{1588--1597}.
\bibitem[{Wang et~al.(2018)Wang, Girshick, Gupta and He}]{wang2018non-local}
\bibinfo{author}{Wang, X.}, \bibinfo{author}{Girshick, R.B.}, \bibinfo{author}{Gupta, A.}, \bibinfo{author}{He, K.}, \bibinfo{year}{2018}.
\newblock \bibinfo{title}{Non-local neural networks}, in: \bibinfo{booktitle}{2018 {IEEE} Conference on Computer Vision and Pattern Recognition}, pp. \bibinfo{pages}{7794--7803}.
\bibitem[{Wang et~al.(2019)Wang, Sun, Liu, Sarma, Bronstein and Solomon}]{dgcnn}
\bibinfo{author}{Wang, Y.}, \bibinfo{author}{Sun, Y.}, \bibinfo{author}{Liu, Z.}, \bibinfo{author}{Sarma, S.E.}, \bibinfo{author}{Bronstein, M.M.}, \bibinfo{author}{Solomon, J.M.}, \bibinfo{year}{2019}.
\newblock \bibinfo{title}{Dynamic graph {CNN} for learning on point clouds}.
\newblock \bibinfo{journal}{{ACM} Trans. Graph.} \bibinfo{volume}{38}, \bibinfo{pages}{146:1--146:12}.
\bibitem[{Wu et~al.(2019)Wu, Qi and Li}]{Pointconv}
\bibinfo{author}{Wu, W.}, \bibinfo{author}{Qi, Z.}, \bibinfo{author}{Li, F.}, \bibinfo{year}{2019}.
\newblock \bibinfo{title}{Pointconv: Deep convolutional networks on 3d point clouds}, in: \bibinfo{booktitle}{{IEEE} Conference on Computer Vision and Pattern Recognition, {CVPR} 2019, Long Beach, CA, USA, June 16-20, 2019}, pp. \bibinfo{pages}{9621--9630}.
\bibitem[{Wu et~al.(2015)Wu, Song, Khosla, Yu, Zhang, Tang and Xiao}]{ModelNet40}
\bibinfo{author}{Wu, Z.}, \bibinfo{author}{Song, S.}, \bibinfo{author}{Khosla, A.}, \bibinfo{author}{Yu, F.}, \bibinfo{author}{Zhang, L.}, \bibinfo{author}{Tang, X.}, \bibinfo{author}{Xiao, J.}, \bibinfo{year}{2015}.
\newblock \bibinfo{title}{3d shapenets: {A} deep representation for volumetric shapes}, in: \bibinfo{booktitle}{{IEEE} Conference on Computer Vision and Pattern Recognition}, pp. \bibinfo{pages}{1912--1920}.
\bibitem[{Xu et~al.(2021)Xu, Ding, Zhao and Qi}]{paconv}
\bibinfo{author}{Xu, M.}, \bibinfo{author}{Ding, R.}, \bibinfo{author}{Zhao, H.}, \bibinfo{author}{Qi, X.}, \bibinfo{year}{2021}.
\newblock \bibinfo{title}{Paconv: Position adaptive convolution with dynamic kernel assembling on point clouds}, in: \bibinfo{booktitle}{{IEEE} Conference on Computer Vision and Pattern Recognition, {CVPR} 2021, virtual, June 19-25, 2021}, pp. \bibinfo{pages}{3173--3182}.
\bibitem[{Xu et~al.(2018)Xu, Fan, Xu, Zeng and Qiao}]{SpiderCNN}
\bibinfo{author}{Xu, Y.}, \bibinfo{author}{Fan, T.}, \bibinfo{author}{Xu, M.}, \bibinfo{author}{Zeng, L.}, \bibinfo{author}{Qiao, Y.}, \bibinfo{year}{2018}.
\newblock \bibinfo{title}{Spidercnn: Deep learning on point sets with parameterized convolutional filters}, in: \bibinfo{booktitle}{Computer Vision - {ECCV} 2018 - 15th European Conference}, pp. \bibinfo{pages}{90--105}.
\bibitem[{Yan et~al.(2018)Yan, Mao and Li}]{second}
\bibinfo{author}{Yan, Y.}, \bibinfo{author}{Mao, Y.}, \bibinfo{author}{Li, B.}, \bibinfo{year}{2018}.
\newblock \bibinfo{title}{Second: Sparsely embedded convolutional detection}.
\newblock \bibinfo{journal}{Sensors} \bibinfo{volume}{18}, \bibinfo{pages}{3337}.
\bibitem[{Yang et~al.(2020)Yang, Sun, Liu and Jia}]{3DSSD}
\bibinfo{author}{Yang, Z.}, \bibinfo{author}{Sun, Y.}, \bibinfo{author}{Liu, S.}, \bibinfo{author}{Jia, J.}, \bibinfo{year}{2020}.
\newblock \bibinfo{title}{3dssd: Point-based 3d single stage object detector}, in: \bibinfo{booktitle}{{IEEE} Conference on Computer Vision and Pattern Recognition}, pp. \bibinfo{pages}{11037--11045}.
\bibitem[{Yang et~al.(2019)Yang, Sun, Liu, Shen and Jia}]{std}
\bibinfo{author}{Yang, Z.}, \bibinfo{author}{Sun, Y.}, \bibinfo{author}{Liu, S.}, \bibinfo{author}{Shen, X.}, \bibinfo{author}{Jia, J.}, \bibinfo{year}{2019}.
\newblock \bibinfo{title}{{STD:} sparse-to-dense 3d object detector for point cloud}, in: \bibinfo{booktitle}{2019 {IEEE/CVF} International Conference on Computer Vision}, pp. \bibinfo{pages}{1951--1960}.
\bibitem[{Yi et~al.(2016)Yi, Kim, Ceylan, Shen, Yan, Su, Lu, Huang, Sheffer and Guibas}]{ShapeNetPart}
\bibinfo{author}{Yi, L.}, \bibinfo{author}{Kim, V.G.}, \bibinfo{author}{Ceylan, D.}, \bibinfo{author}{Shen, I.}, \bibinfo{author}{Yan, M.}, \bibinfo{author}{Su, H.}, \bibinfo{author}{Lu, C.}, \bibinfo{author}{Huang, Q.}, \bibinfo{author}{Sheffer, A.}, \bibinfo{author}{Guibas, L.J.}, \bibinfo{year}{2016}.
\newblock \bibinfo{title}{A scalable active framework for region annotation in 3d shape collections}.
\newblock \bibinfo{journal}{{ACM} Trans. Graph.} \bibinfo{volume}{35}, \bibinfo{pages}{210:1--210:12}.
\bibitem[{Zhang et~al.(2023)Zhang, Wang, Wang, Gao, Li and Shi}]{pointnn}
\bibinfo{author}{Zhang, R.}, \bibinfo{author}{Wang, L.}, \bibinfo{author}{Wang, Y.}, \bibinfo{author}{Gao, P.}, \bibinfo{author}{Li, H.}, \bibinfo{author}{Shi, J.}, \bibinfo{year}{2023}.
\newblock \bibinfo{title}{Parameter is not all you need: Starting from non-parametric networks for 3d point cloud analysis}.
\newblock \bibinfo{journal}{arXiv preprint arXiv:2303.08134} .
\bibitem[{Zhang et~al.(2019)Zhang, Zhou, Sun, Wang, Shi and Loy}]{mmmot}
\bibinfo{author}{Zhang, W.}, \bibinfo{author}{Zhou, H.}, \bibinfo{author}{Sun, S.}, \bibinfo{author}{Wang, Z.}, \bibinfo{author}{Shi, J.}, \bibinfo{author}{Loy, C.C.}, \bibinfo{year}{2019}.
\newblock \bibinfo{title}{Robust multi-modality multi-object tracking}, in: \bibinfo{booktitle}{2019 {IEEE/CVF} International Conference on Computer Vision}, pp. \bibinfo{pages}{2365--2374}.
\bibitem[{Zhang et~al.(2021)Zhang, Huang and Wang}]{pc-rgnn}
\bibinfo{author}{Zhang, Y.}, \bibinfo{author}{Huang, D.}, \bibinfo{author}{Wang, Y.}, \bibinfo{year}{2021}.
\newblock \bibinfo{title}{Pc-rgnn: Point cloud completion and graph neural network for 3d object detection}, in: \bibinfo{booktitle}{Proceedings of the AAAI Conference on artificial intelligence}, pp. \bibinfo{pages}{3430--3437}.
\bibitem[{Zhao et~al.(2021)Zhao, Jiang, Jia, Torr and Koltun}]{PointTransformer}
\bibinfo{author}{Zhao, H.}, \bibinfo{author}{Jiang, L.}, \bibinfo{author}{Jia, J.}, \bibinfo{author}{Torr, P.H.S.}, \bibinfo{author}{Koltun, V.}, \bibinfo{year}{2021}.
\newblock \bibinfo{title}{Point transformer}, in: \bibinfo{booktitle}{2021 {IEEE/CVF} International Conference on Computer Vision}, pp. \bibinfo{pages}{16239--16248}.

\end{thebibliography}

\end{document}